\documentclass[runningheads]{llncs}
\usepackage{graphicx}
\usepackage{amsmath,amssymb} % define this before the line numbering.
\usepackage{color}
\usepackage{times}
\usepackage{epsfig}
\usepackage{graphicx}
\usepackage{amsmath}
\usepackage{amssymb}
\usepackage[tight,scriptsize]{subfigure}
\usepackage[capitalize]{cleveref}

\def\etal{\emph{et al}.}

\newcommand{\pyramid}{Bottom-Heavy-I3D}
\newcommand{\invpyramid}{Top-Heavy-I3D}
\newcommand{\spyramid}{Bottom-Heavy-S3D}
\newcommand{\sinvpyramid}{Top-Heavy-S3D}
\newcommand{\Sep}{S3D} % separable
\newcommand{\SG}{S3D-G} % gating

\newcommand{\Sthree}{S3D} % Sep plus gating
\newcommand{\Ithree}{I3D}
\newcommand{\Itwo}{I2D}

\newcommand{\MK}{Mini-Kinetics-200} 
\newcommand{\Something}{Something-something}
\newcommand{\FK}{Kinetics-Full}
\newcommand{\Kinetics}{Kinetics}

\begin{document}
\title{Rethinking Spatiotemporal Feature Learning: Speed-Accuracy Trade-offs in Video Classification}
% Replace with your title

\titlerunning{Rethinking Spatiotemporal Feature Learning}% Replace with a meaningful short version of your title
\author{Saining Xie\inst{1,2}\and
Chen Sun\inst{1} \and
Jonathan Huang \inst{1} \and
Zhuowen Tu \inst{1,2} \and
Kevin Murphy \inst{1} 
}
%
%Please write out author names in full in the paper, i.e. full given and family names. 
%If any authors have names that can be parsed into FirstName LastName in multiple ways, please include the correct parsing, in a comment to the volume editors:
%\index{Lastnames, Firstnames}
%(Do not uncomment it, because you may introduce extra index items if you do that, we will use scripts for introducing index entries...)
\authorrunning{Saining Xie \emph{et al}.}
% Replace with shorter version of the author list. If there are more authors than fits a line, please use A. Author et al.
%

\institute{Google Research\and University of California San Diego}
\maketitle              % typeset the header of the contribution

\begin{abstract}
Despite the steady progress in video analysis led by the adoption of convolutional neural networks (CNNs), the relative improvement has been less drastic as that in 2D static image classification. Three main challenges exist including spatial (image) feature representation, temporal information representation, and model/computation complexity. It was recently shown by Carreira and Zisserman that 3D CNNs, inflated from 2D networks and pretrained on ImageNet, could be a promising way for spatial and temporal representation learning. However, as for model/computation complexity, 3D CNNs are much more expensive than 2D CNNs and prone to overfit. We seek a balance between speed and accuracy by building an effective and efficient video classification system through systematic exploration of critical network design choices. 
In particular, we show that it is possible to replace many of the 3D convolutions by low-cost 2D convolutions. Rather surprisingly, best result (in both speed and accuracy) is achieved when replacing the 3D convolutions at the bottom of the network, suggesting that temporal representation learning on high-level ``semantic'' features is more useful. Our conclusion generalizes to datasets with very different properties. When combined with several other cost-effective designs including separable spatial/temporal convolution and feature gating, our system results in an effective video classification system that that produces very competitive results on several action classification benchmarks (Kinetics, Something-something, UCF101 and HMDB), as well as two action detection (localization) benchmarks (JHMDB and UCF101-24).
\end{abstract}

\section{Introduction}

The resurgence of convolutional neural networks (CNNs) has led to a wave of unprecedented advances for image classification using end-to-end hierarchical feature learning architectures \cite{krizhevsky2012imagenet,InceptionV1,simonyan2015very,resnet}. The task of video classification, however, has not enjoyed the same level of performance jump as in image classification.
In the past, one limitation was the lack of large-scale labeled video datasets.
However, the recent creation of
Sports-1M \cite{Karpathy2014},
Kinetics \cite{kay2017kinetics},
\Something\ \cite{Something},
ActivityNet \cite{ActivityNet},
Charades \cite{Charades}, etc.
has partially removed that impediment.

Now we face more fundamental challenges.
In particular, we have three main barriers to overcome: (1) 
how best to represent  spatial information (i.e., recognizing the appearances of objects);
(2) how best to represent temporal information
(i.e., recognizing  context, correlation and causation through time);
and (3) 
how best to tradeoff model complexity with speed,
 both at training and testing time.

\begin{figure}[t]
\centering
\includegraphics[height=2.2in]{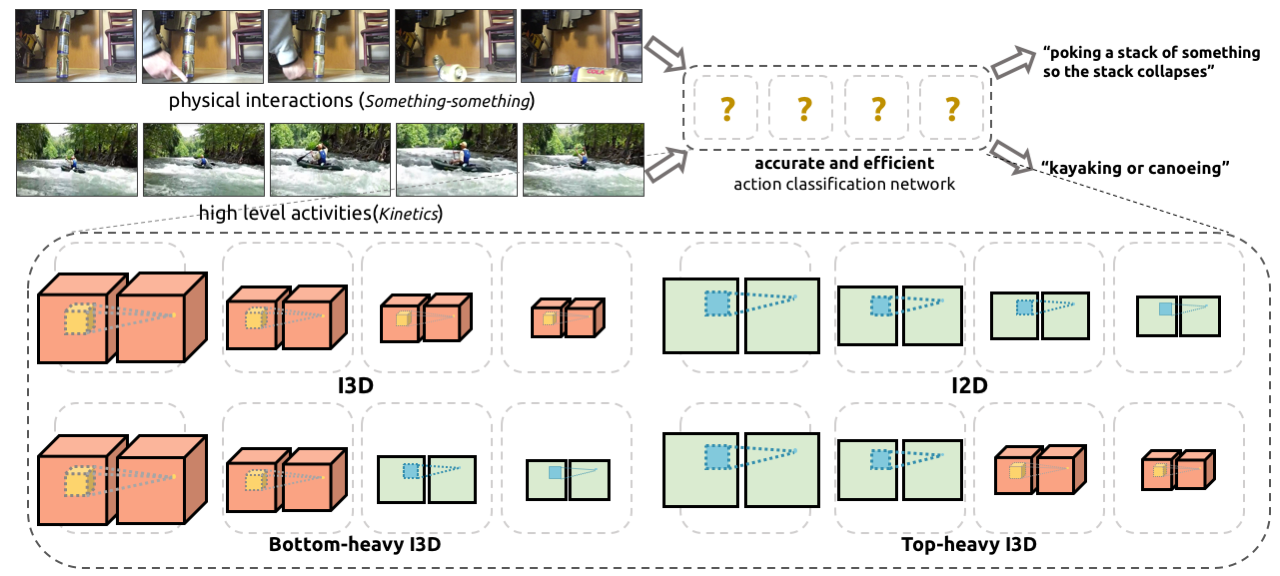}
\caption{
Our goal is to classify videos into different categories, as shown in the top row.
We focus on two qualitatively different kinds of datasets:
\Something\, which requires recognizing low-level physical interactions,
and Kinetics, which requires recognizing high-level activities.
The main question we seek to answer is what kind of network architecture to use.
We consider 4 main variants:
I2D, which is a 2D CNN, operating on multiple frames;
I3D, which is a 3D CNN, convolving over space and time;
Bottom-Heavy I3D, which uses 3D in the lower layers, and 2D in the higher layers;
and 
Top-Heavy I3D, which uses 2D in the lower (larger) layers,
and 3D in the upper layers.
}
\label{fig:teaser}
\end{figure}

In this paper, we study these three questions by considering various kinds of 3D CNNs.
Our starting point is the state of the art approach, due to
Carreira and Zisserman~\cite{carreira2017quo},
known as ``I3D'' 
(since it ``inflates'' the 2D convolutional filters of the
``Inception'' network \cite{InceptionV1} to 3D).

Despite giving good performance, 
this model is very computationally expensive.  This prompts several questions, which we seek to 
address in this paper:
\begin{itemize}

\item Do we even need 3D convolution? If so, what layers should we make 3D, and what layers can be 2D? Does this depend on the nature of the dataset and task?

\item Is it important that we convolve jointly over time and space, or would it suffice to convolve over these dimensions independently?

\item How can we use answers to the above questions to improve on prior methods in terms of accuracy, speed and memory footprint?
\end{itemize}

To answer the first question,
we apply ``network surgery'' to obtain several
variants of the I3D architecture.
In one family of variants, which we call \pyramid, we retain 3D temporal convolutions at the lowest layers of the network
(the ones closest to the pixels), and use 2D convolutions for the higher layers.
In the other family of variants, which we call \invpyramid, we do the opposite, and retain 3D temporal convolutions at the top layers, and use 2D for the lower layers
(see Figure~\ref{fig:teaser}).
We then investigate how to trade between accuracy and speed by varying 
the number of layers that are ``deflated'' (converted to 2D) in this way. 
We find that the \invpyramid\ models are faster,
which is not surprising, since they only apply 3D to the abstract feature maps, which are smaller than the low level feature maps due to spatial pooling.
However, we also find that \invpyramid\ models are often more accurate, which is surprising since they ignore low-level motion cues.

To answer the second question (about separating space and time), we consider replacing 3D convolutions
with spatial and temporal separable 3D convolutions,
i.e., we replace filters of the form $k_t \times k \times k$
by $1 \times k \times k$ followed by $k_t \times 1 \times 1$,
where $k_t$ is the width of the filter in time,
and $k$ is the height/width of the filter in space.
We call the resulting model \Sep, which stands for ``separable 3D CNN''.
\Sep\ obviously has many fewer parameters than models that use standard 3D convolution,
and it is more computationally efficient.
Surprisingly, we also show that it also has better accuracy than the original I3D model.

Finally, to answer the third question (about putting things together for an efficient and accurate video classification system), we combine what we have learned in answering
the above two questions with a spatio-temporal gating mechanism
to design a new model architecture
which we call \SG.
We show that this model gives significant gains in accuracy over baseline methods on  a variety of challenging video classification datasets, such as Kinetics,
\Something, UCF-101 and HMDB,
and also outperforms many other methods on other video recognition tasks, such as action localization on JHMDB.

\section{Related work}
\label{sec:related}

2D CNNs have achieved state of the art results for image classification,
so, not surprisingly, 
 there have been many recent attempts to extend these successes
to video classification.
 The Inception 3D (I3D) architecture~\cite{carreira2017quo} proposed by Carreira and Zisserman is one of the current
 state-of-the-art models. There are three key ingredients for its success: first, they ``inflate'' all the 2D convolution filters used by the Inception V1 architecture~\cite{InceptionV1} into 3D convolutions, and carefully choose the temporal kernel size in the earlier layers. Second, they initialize the inflated model weights by duplicating weights
 that were pre-trained on ImageNet classification
 over the temporal dimension. Finally, they train the network on
  the large-scale Kinetics dataset~\cite{kay2017kinetics}.

Unfortunately, 3D CNNs are computationally expensive, so there has been recent interest in more efficient variants.
In concurrent work, \cite{Tran2018}
has recently proposed a variety of models based on top
of the ResNet architecture \cite{resnet}.
In particular, they consider models that use 3D convolution in either the bottom or top layers, and 2D in the rest; they call these ``mixed convolutional'' models.
This is similar to our top-heavy and bottom-heavy models.
They conclude that bottom heavy networks are more accurate,
which contradicts our finding.
However, the differences they find between top heavy and bottom heavy are fairly small,
and are conflated with changes in computational complexity.
By studying the entire speed-accuracy tradeoff curve (of Inception variants),
we show that there are clear benefits to using a top-heavy design for a given computational budget (see Section~\ref{sec:some3d}).

Another way to save computation is to replace 3D convolutions with separable convolutions, in which we first convolve spatially in 2D, and then convolve temporally in 1D.
We call the resulting model S3D.
This factorization is similar in spirit to the depth-wise separable convolutions
 used in \cite{xception,mobilenet,resnext}, except that we apply the idea
 to the temporal dimension instead of the feature dimension.
This idea has been used in a variety of recent papers,
including \cite{Tran2018} (who call it ``R(2+1)D''),
\cite{P3D} (who call it ``Pseudo-3D network''),
\cite{Sun2015} (who call it ``factorized spatio-temporal convolutional networks''),
etc.
We use the same method, but combine it with both top-heavy and bottom-heavy designs, which is a combination that leads to a very efficient video classification system.
We show that the gains from separable convolution are complementary to the gains from using a top-heavy design (see Section~\ref{sec:sepconv}).

An efficient way to improve accuracy is to use feature gating, which captures dependencies between feature channels with a simple but effective multiplicative transformation.
This can be viewed as an efficient approximation to second-order pooling
as shown in \cite{Girdhar_17b_AttentionalPoolingAction}.
Feature gating has been used for many tasks, 
such as machine translation~\cite{dauphin2016language}, 
VQA~\cite{perez2017film}, 
reinforcement learning~\cite{elfwing2018sigmoid}, image classification~\cite{ramachandran2017swish,hu2017squeeze},
and action recognition~\cite{miech2017learnable}.
We consider a variant of the above techniques in 
which we place the feature gating module after each of the temporal convolutions in an S3D network,
and show that this results in substantial gains in accuracy (see Section~\ref{sec:gating}).

Another way to improve accuracy (at somewhat higher cost)
is to use precomputed optical flow features.
This idea was successfully used in \cite{TwoStreamVGG},
who proposed a two-stream architecture where one CNN stream handles raw RGB input,  and the other
 handles  precomputed optical flow. 
Since then, many video classification methods follow the same multi-stream 2D CNN design, and have made improvements in terms of new representations~\cite{bilen2016dynamic,bilen2017action}, different backbone architecture~\cite{FeichtenhoferCVPR17,TwoStreamCUHK,ng_cvpr15,Girdhar_17b_AttentionalPoolingAction}, fusion of the streams~\cite{Feichtenhofer16,feichtenhofer2017,FeichtenhoferNIPS16,chained_fusion} and exploiting richer temporal structures~\cite{lrcn2014,Wang_Transformation,tsn_wang_eccv16}.
We will study the benefits of using optical flow in Section~\ref{sec:flow}.

\section{Experiment Setup}
\label{sec:experiments}
\subsection{Datasets}
\label{data}

In this paper, we consider two large video action classification datasets.
The first one is Kinetics \cite{kay2017kinetics},
  which  is a large dataset collected from YouTube,  containing 400 action classes and 240K training examples.
  Each example is temporally trimmed to be around 10 seconds.
  Since the full Kinetics dataset is quite large, 
  we have created a smaller dataset
that we call \MK.\footnote{
The original ``Mini-Kinetics'' dataset used in 
 \cite{kay2017kinetics} contains videos that are no longer available.
 We created the new \MK\ dataset in collaboration with the original authors.
} %
\MK\ consists of 
the 200 categories with most training examples;
for each category, we randomly sample 400 examples from the training set, and 25 examples from the validation set,
resulting in
80K training examples and 5K validation examples in total. The splits are publicly released to enable future comparisons.
We also report some results on the original Kinetics dataset, which we will call \FK\ for clarity.

The second main dataset is
\Something~\cite{Something}. It consists of ~110k videos of 174 different low-level actions, each lasting  between 2 to 6 seconds.
In contrast to Kinetics,
this dataset requires making fine-grained low-level distinctions, such as between ``Pushing something from left to right'' and ``Pushing something from right to left''.
It is therefore an interesting question whether the same principles will hold and the same architectures will work well on both datasets.

We also consider two 
smaller action classification datasets to test the transferability of our model, 
which we discuss in Section~\ref{sec:otherClassification},
as well as two 
action detection datasets, which we discuss in Section~\ref{sec:detection}.

\subsection{Model training}

Our training procedure largely follows~\cite{carreira2017quo}. During training, we densely sample 64 frames from a video and resize input frames to $256\times 256$ and then take random crops of size $224\times 224$. During evaluation, we use all frames and take $224\times 224$ center crops from the resized frames. Our models are implemented with TensorFlow and optimized with a vanilla synchronous SGD algorithm with momentum of 0.9 and on 56 GPUs, batch size is set to 6 per GPU. For \MK, we train our model for 80k steps with an initial learning rate of 0.1. We decay the learning rate at step 60k to 0.01, and step 70k to 0.001. Since \Something\ is a smaller dataset, we reduce the number of GPUs to 16 and train at learning rate of 0.1 for 10k steps.

\subsection{Measuring speed and accuracy}

We report top-1 and top-5 accuracy. To measure the computational efficiency of our models, we report theoretical FLOPS based on a single input video sequence of 64 frames and spatial size $224\times 224$. We pad the total number of frames to 250 for \MK\ and 64 for \Something\ when evaluating.
\section{Network surgery}
\label{sec:surgery}

In this section, we report the results of various ``network surgery'' experiments, where we vary different aspects of the I3D model to study the effects on speed and accuracy.

\begin{figure}[!htp]
\begin{center}
\subfigure[I3D]{
\includegraphics[height=25mm]{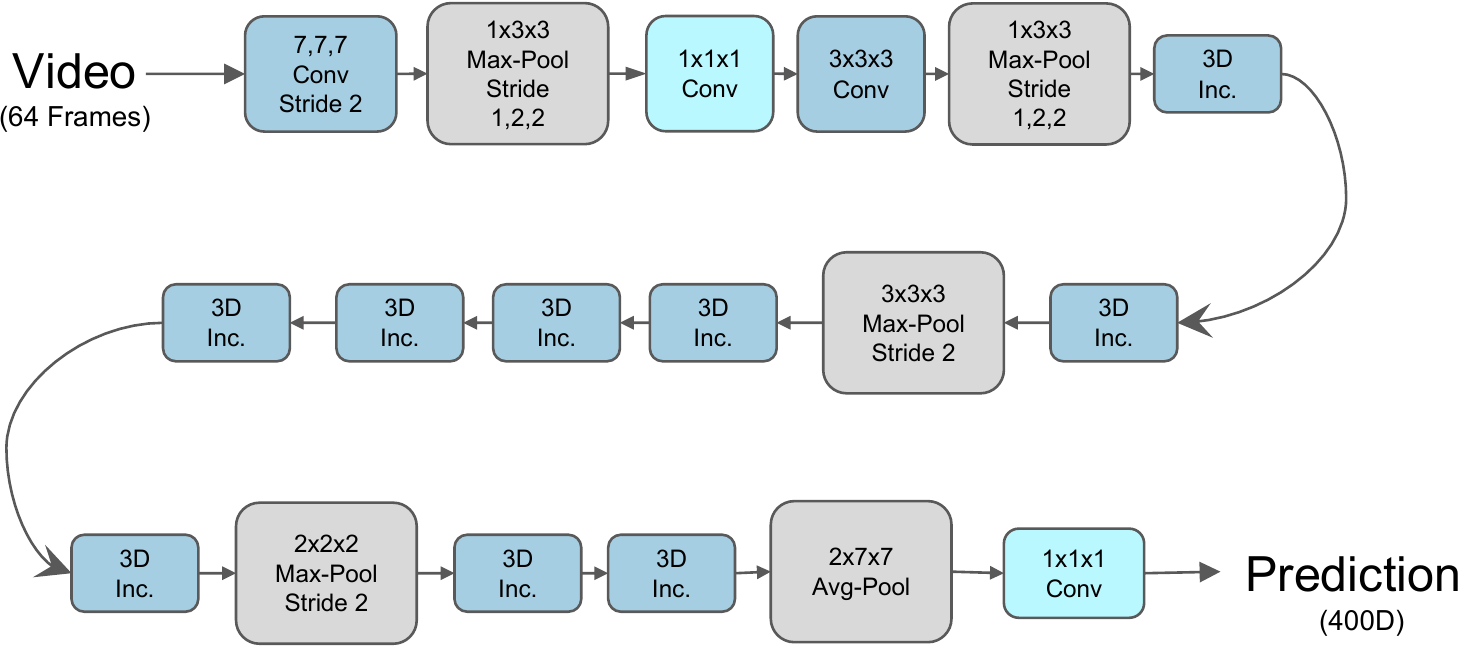}
\label{fig:detail_i3d}
}
\subfigure[I2D]{
\includegraphics[height=25mm]{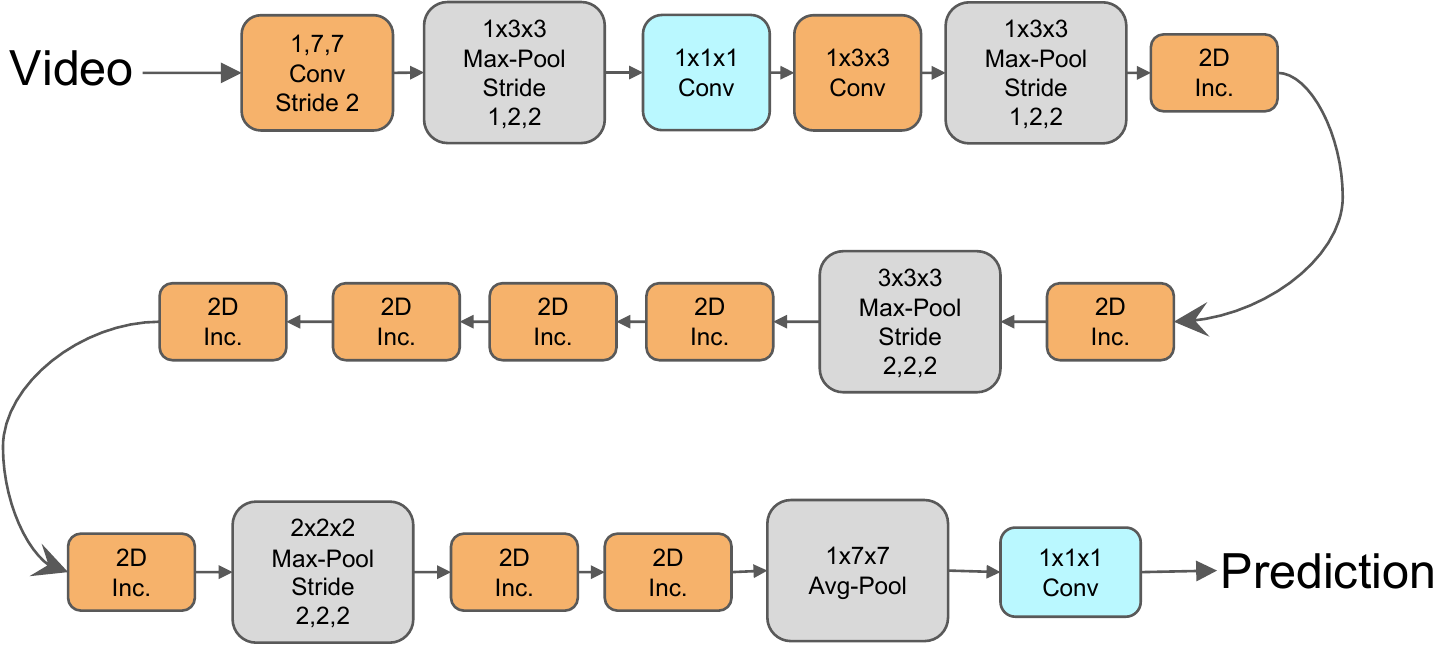}
\label{fig:detail_i2d}
}
\subfigure[Bottom-heavy I3D]{
\includegraphics[height=25mm]{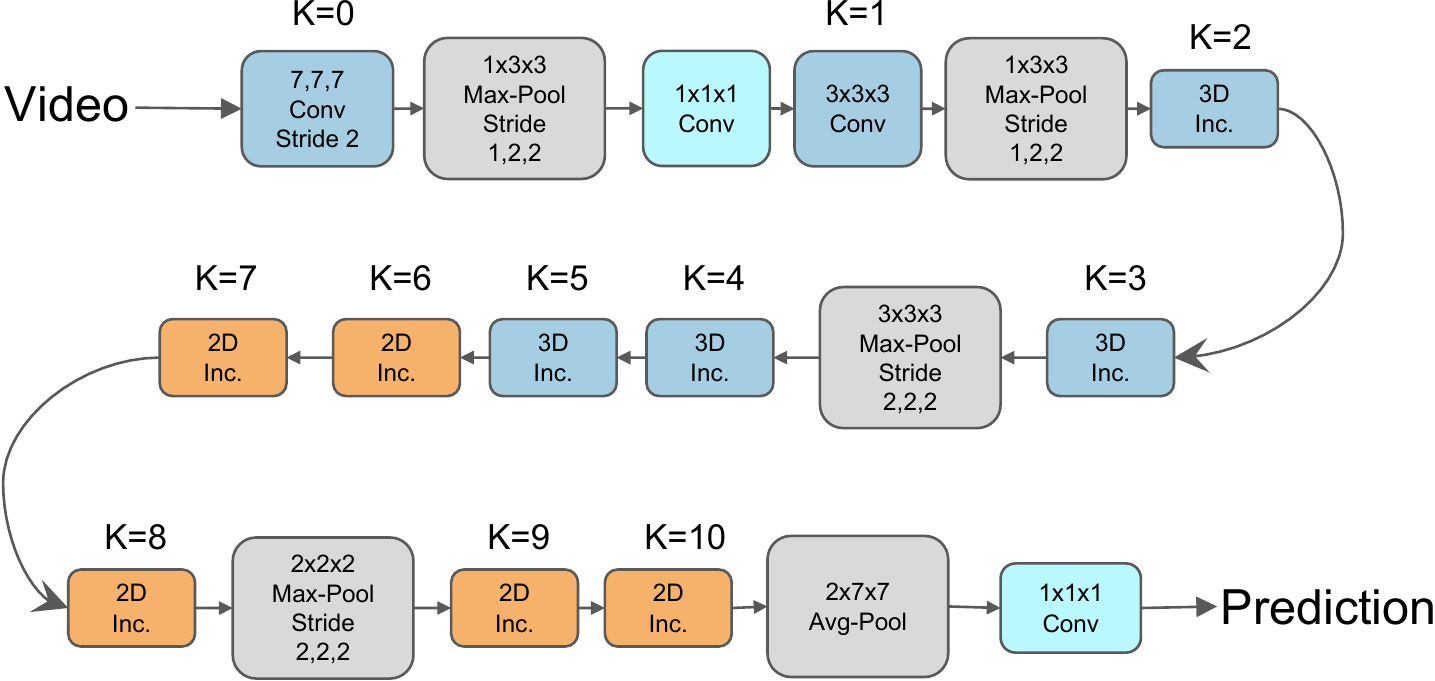}
\label{fig:detail_bottomheavy_i3d}
}
\subfigure[Top-heavy I3D]{
\includegraphics[height=25mm]{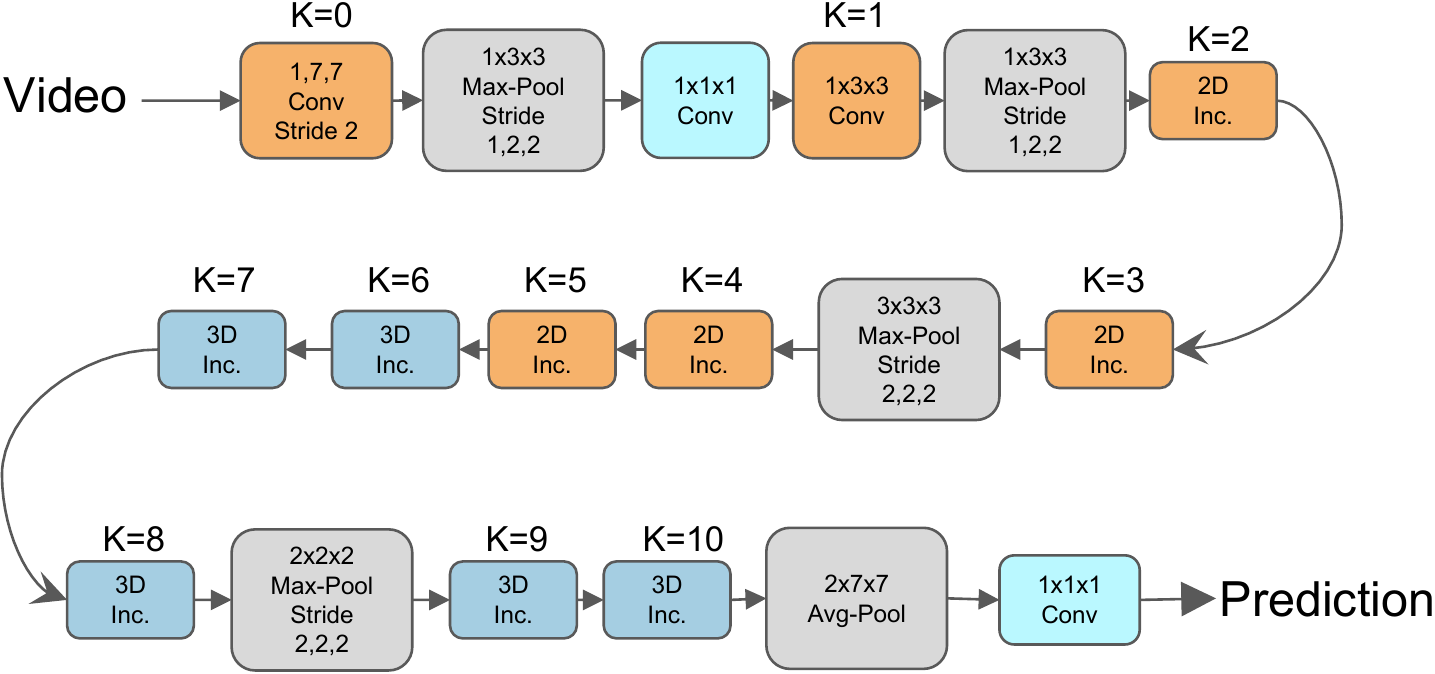}
\label{fig:detail_topheavy_i3d}
}
\end{center}
\caption{
Network architecture details for (a) I3D, (b) I2D, (c) Bottom-Heavy and (d) Top-Heavy variants. $K$ indexes the spatio-temporal convolutional layers. The ``2D Inc.'' and ``3D Inc.'' blocks
refer to 2D and 3D inception blocks, defined in Figure~\ref{fig:inc_blocks}.}
\end{figure}

\begin{figure}
\begin{center}
\subfigure[]{
\includegraphics[height=35mm]{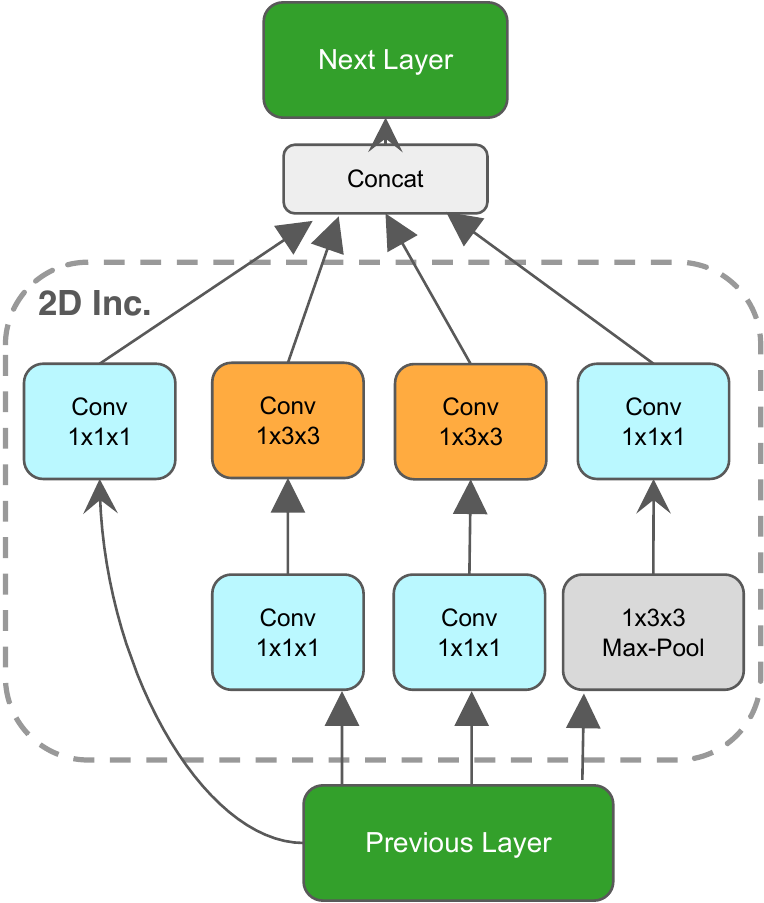}
\label{fig:inc_block_2d}
}\qquad
\subfigure[]{
\includegraphics[height=35mm]{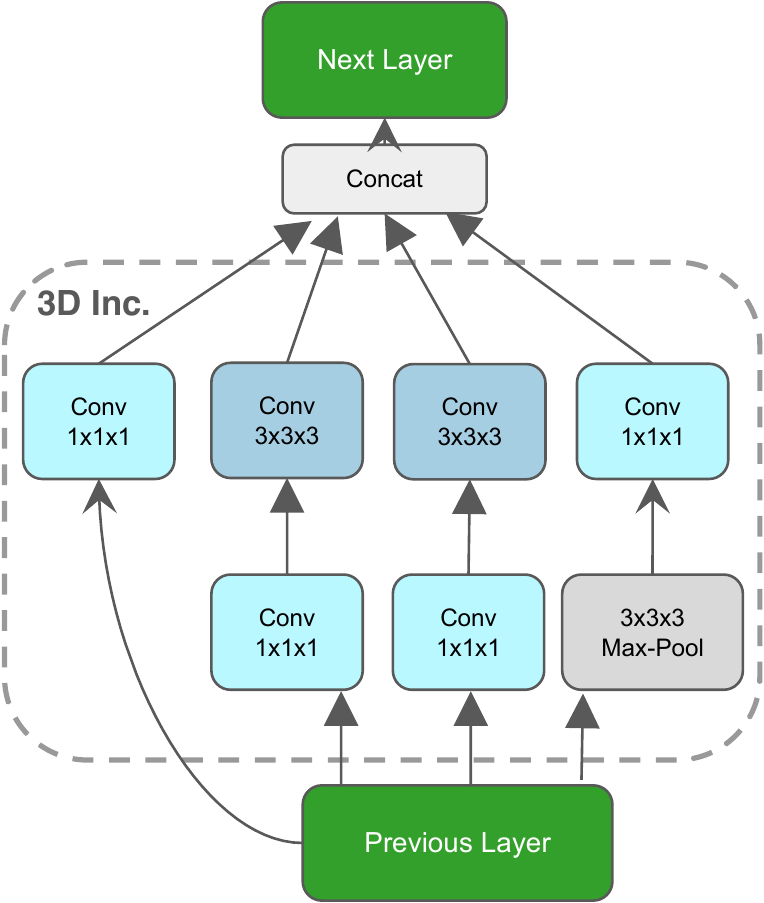}
\label{fig:inc_block_3d}
}\qquad
\subfigure[]{
\includegraphics[height=35mm]{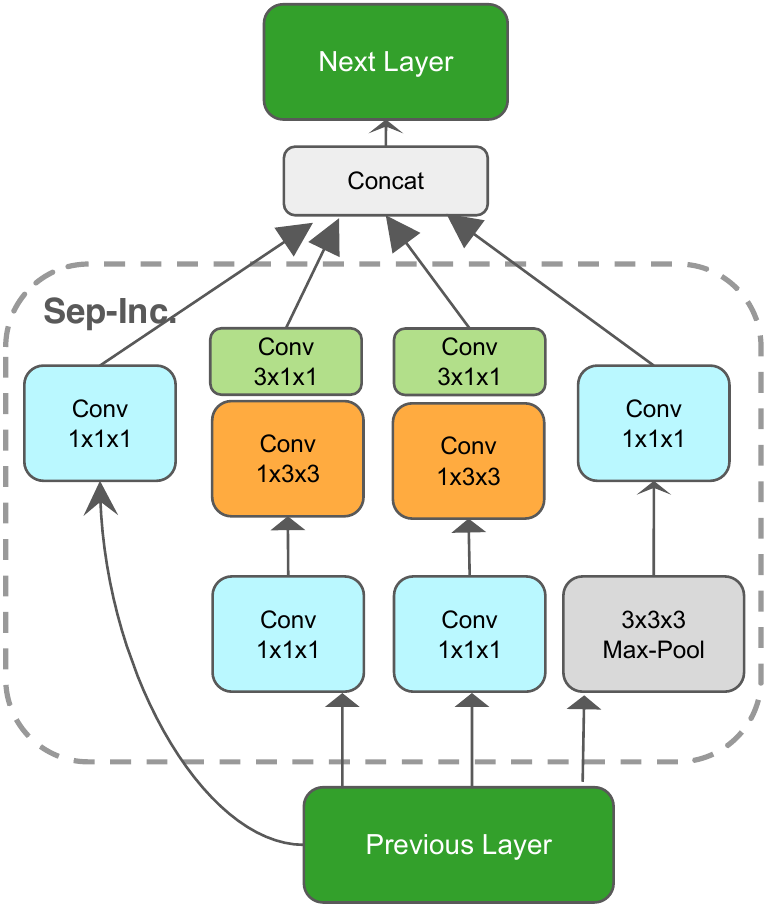}
\label{fig:sep_inc_block}
}
\end{center}
\caption{\subref{fig:inc_block_2d} 2D Inception block;
\subref{fig:inc_block_3d} 3D Inception block;
\subref{fig:sep_inc_block} 3D temporal separable Inception block used in S3D networks.
}
\label{fig:inc_blocks}
\end{figure}

\subsection{Replacing \emph{all} 3D convolutions with 2D}
\label{sec:all3d}
\label{sec:pyramids}

In this section, we seek to determine how much value  3D convolution brings,
motivated by the surprising success of 2D CNN approaches
to video classification (see e.g., \cite{tsn_wang_eccv16}).
We do this by 
replacing every 3D filter in the I3D model with a 2D filter.  This yields what we will refer to as the I2D
model.\footnote{
To reduce the memory and time requirements,
and to keep
the training protocol identical to I3D (in terms of the number of clips we use for training in each batch, etc),
we retain two max-pooling layers with temporal stride 2 between 
Inception modules.
Hence, strictly speaking, I2D is not a pure 2D model. However, it is very similar to a single-frame 2D classification model.
}

Theoretically, the I2D network should be invariant to the temporal reversal of the input frames, since it is not capable of incorporating global signals. To verify this, we train I2D and the original I3D model on the \FK\ and \Something\ datasets with normal frame order, and apply the trained models on validation data in which the frames are in
normal order and reversed temporal order.
The results of the experiment are shown in Table~\ref{tab:arrowOfTime}.
We see that I2D has the same performance on both versions during testing, as is to be expected. However, 
we notice an interesting difference between the \Kinetics\ dataset and the \Something\ dataset.
In the former case, the performance of I3D is indifferent to the  ``arrow of time''~\cite{pickup2014seeing},
whereas in the latter case, reversing the order hurts performance.
We believe this is because \Something\ dataset requires fine-grained distinctions between visually similar action categories.

\setlength{\tabcolsep}{3pt}
\begin{table}
\begin{center}
\small
\begin{tabular}{c|c|c|c|c}
\hline
& \multicolumn{2}{c|}{\FK} & \multicolumn{2}{c}{\Something} \\
\hline
 Model  & Normal  (\%) & Reversed (\%) & Normal  (\%) & Reversed (\%)\\
\hline
\Ithree   & 71.1 & 71.1 & 45.8 & 15.2\\
\Itwo   & 67.0 & 67.2 & 34.4 & 35.2\\
\hline
\end{tabular}
\end{center}
\caption{Top-1 accuracy on \FK\ and \Something\ datasets. 
We train on frames in normal order, and then test on frames
in normal order or reverse order.
Not surprisingly, 2D CNNs do not care about the order of the frames.
For 3D CNNs on \FK\, the results are the same on normal and reverse order, 
indicating
that capturing the ``arrow of time'' is not important on this dataset.
However, on \Something\, the exact order does matter.
}
\label{tab:arrowOfTime}
\end{table}

\subsection{Replacing \emph{some} 3D convolutions with 2D}
\label{sec:some3d}

Although we have seen that 3D convolution can boost accuracy compared to 2D convolution, it is computationally very expensive.
In this section, we investigate the consequences of only replacing some of the 3D convolutions with 2D.
Specifically, starting with an I2D model, we gradually inflate 2D convolutions into 3D, 
from low-level to high-level layers in the network,
to create what we call the \pyramid\ model.
We also consider the opposite process, in which we inflate the top layers of the model to 3D, but keep the lower layers 2D;
we call such models \invpyramid\ models.

\begin{figure}
\centering
\begin{tabular}{cc}
\includegraphics[height=1.5in]{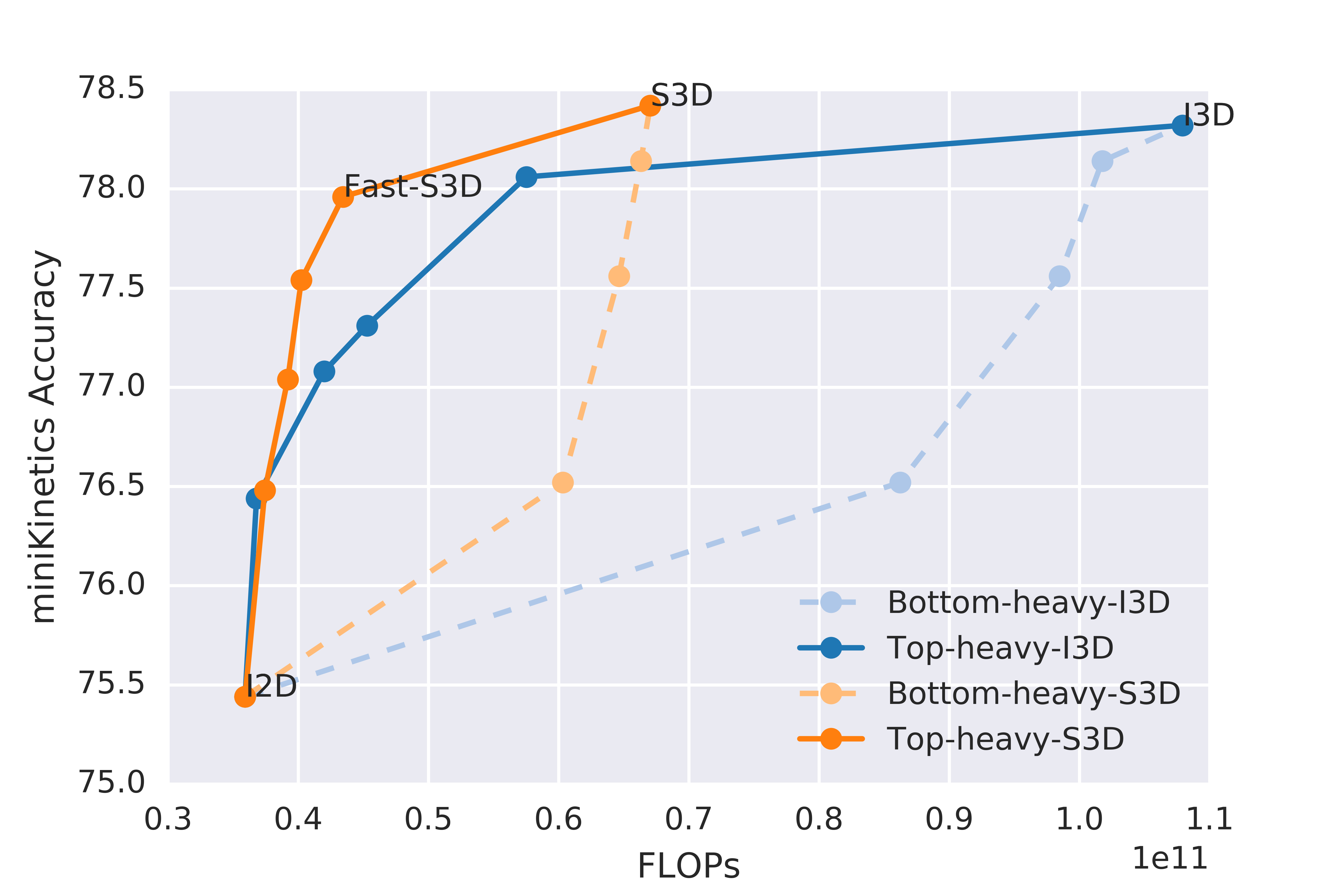}
&
\includegraphics[height=1.5in]{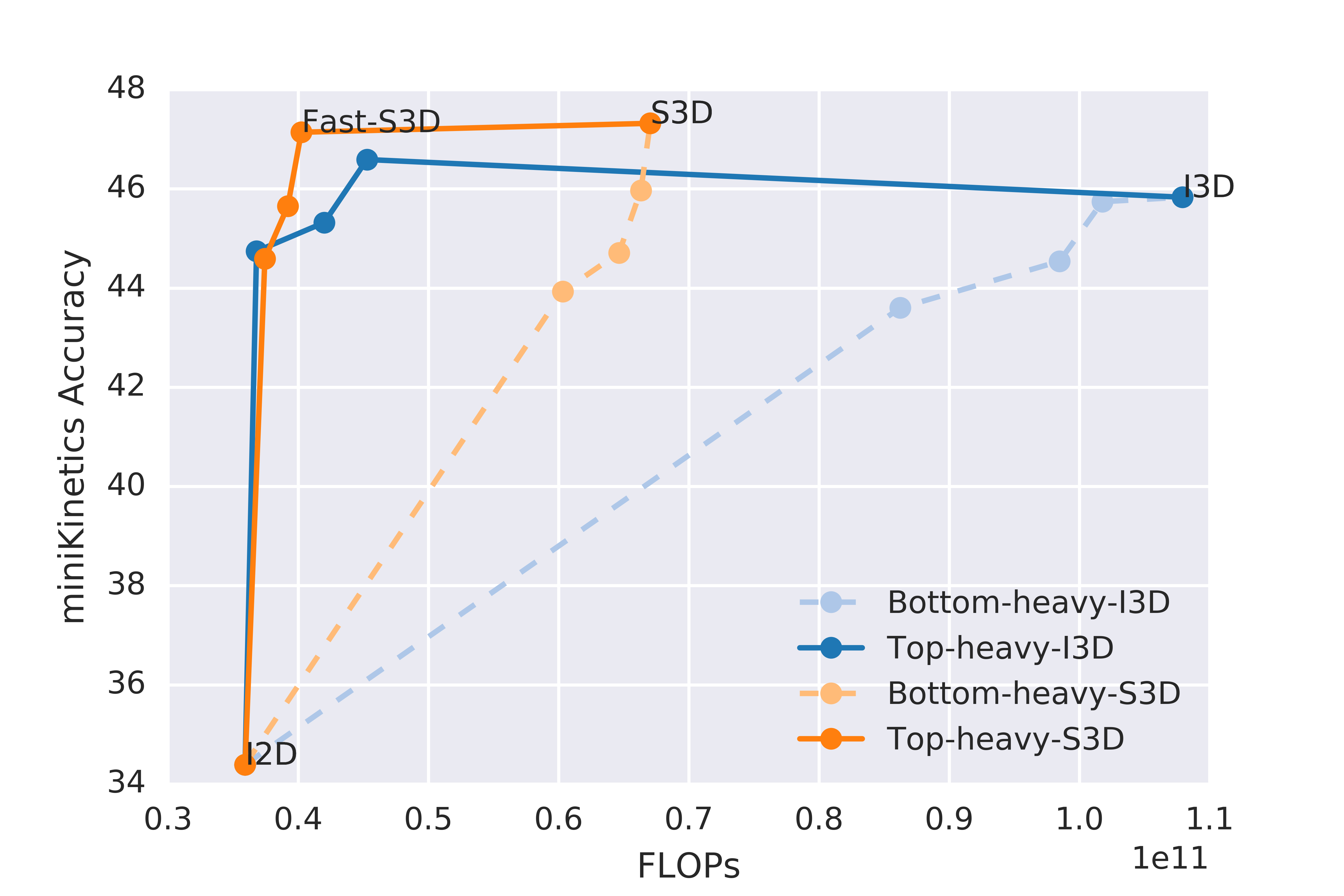}
\\
(a) & (b)
\end{tabular}
\caption{Accuracy vs number of FLOPS needed to perform inference on 64 RGB frames. 
Left: \MK\ dataset.
Right: \Something\ dataset.
Solid lines  denote top-heavy models,
dotted lines denote bottom-heavy models.
Orange denotes spatial and temporal separable 3D convolutions, blue denotes full 3D convolutions.}
\label{fig:accuracyVsFlops}
\end{figure}

We train and evaluate the \pyramid\ and \invpyramid\ models on \MK\ and \Something,
and show the results in Figures~\ref{fig:accuracyVsFlops}.
We see that the solid blue lines (top heavy I3D) are much better than the dotted blue lines (bottom heavy I3D) under the same FLOPS, which indicates that top heavy models are faster and more accurate.
The speed increase is expected, since  in a top-heavy model,
the feature maps are reduced in size using spatial pooling before being convolved in 3D. For fixed computation budget, Top-Heavy-I3D is often significantly more accurate than Bottom-Heavy-I3D. This suggests that 3D convolutions are more capable and useful to model temporal patterns amongst high level features that are rich in semantics.

\begin{figure}[t]
\centering
  \includegraphics[width=12cm,height=2cm]{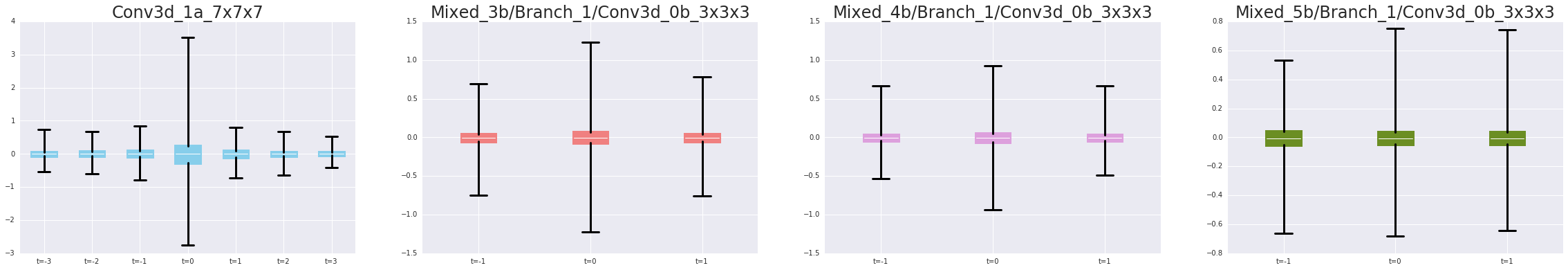}
  \caption{Statistics of convolutional filter weights of an I3D model trained on \FK. Each boxplot shows the distribution of $W_l(t,:,:,:)$ for temporal offset $t$, with $t=0$ being in the middle.
  Results for
  different layers $l$ are shown in different panels,
  with lowest layers on the left.
 All filters with different temporal offset are initialized with the same set of weights. Low-level
 filters essentially ignore the temporal dimension,
 unlike higher level filters,
 where the weights distributed nicely across different temporal offsets. }
  \label{fig:weights}
\end{figure}

\subsection{Analysis of weight distribution of learned filters}

To verify the above intuition, we examined the  weights of an I3D model
which was trained on \FK.
Figure~\ref{fig:weights} shows the distribution of these weights across  4 layers of our model, from low-level to high-level. In particular, each boxplot shows the distribution of $W_l(t,:,:,:)$ for temporal offset $t$ and layer $l$. 
We use $t=0$ to indicate no offset in time, i.e.,  
the center in the temporal kernel. 
At initialization, all the filters started with the same set of (2D convolution) weights (derived from an Inception model pre-trained on Imagenet) for each value of $t \in \{-1,0,1\}.$
After training, we see that the temporally offset filters (i.e., for $t \neq 0$) have a weight distribution that is still closely centered on zero in the lower layers (see left panel), whereas the variance of the distribution increases in higher layers (see right panel).
This suggests once again that the higher level temporal patterns are more useful for the Kinetics action classification task.

\subsection{Separating temporal convolution from spatial convolutions}
\label{sec:sepconv}

\begin{figure}[!htp]
\begin{center}
\includegraphics[height=30mm]{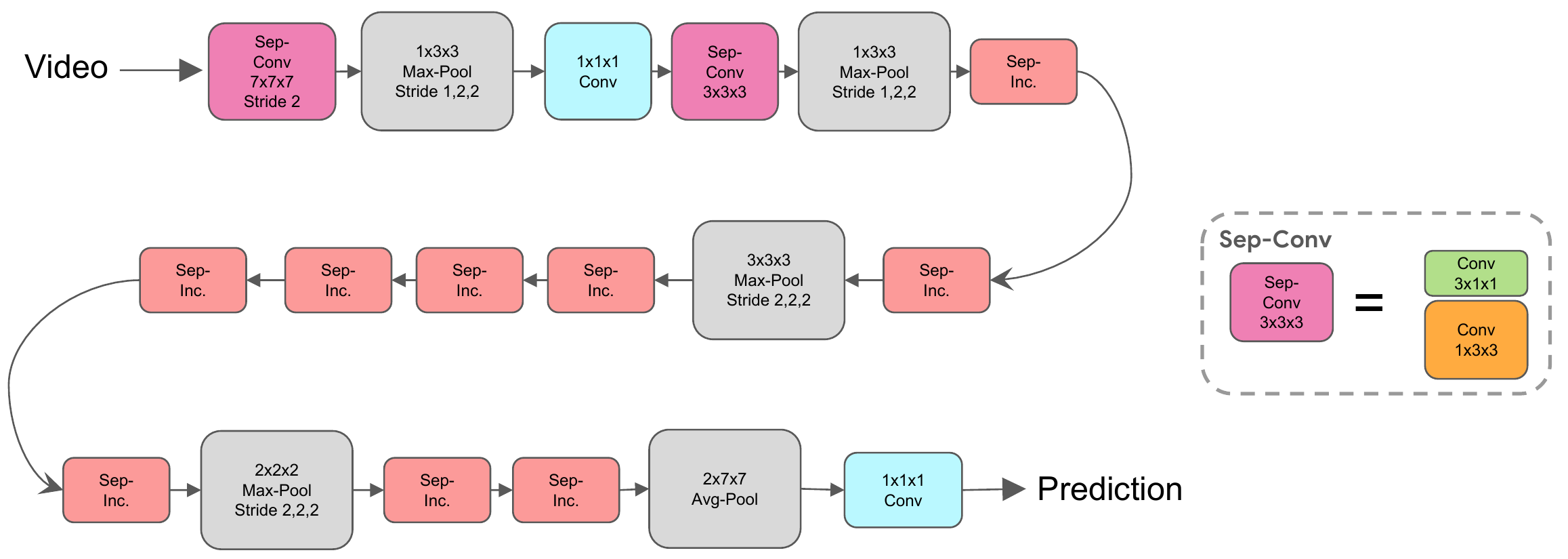}
\end{center}
\caption{
An illustration of the S3D model.
Dark red boxes are
temporal separable convolutions (sep-conv),
and pink boxes are temporal separable
inception blocks, shown in 
Figure~\ref{fig:sep_inc_block}.
}
\label{fig:s3d}
\end{figure}

In this section, we study the effect of replacing  standard 3D convolution with a
factored version which disentangles this operation into a temporal part and a spatial part.
In more detail, our method is to 
 replace each 3D convolution with two consecutive convolution layers: one 2D convolution layer to learn spatial features, followed by a 1D convolution layer purely on the temporal axis.
This can be implemented  by running two 3D convolutions, where the first (spatial) convolution has filter shape $[1,k,k]$ and the 
second (temporal) convolution has filter shape $[k, 1, 1]$.
By applying this factorization to I3D, we obtain a
model which we refer to as S3D. For a detailed illustration of the architecture, please refer to Figure~\ref{fig:s3d}.\footnote{
There are 4 branches in an Inception block, but only two of them have 3x3 convolutions (the other two being pointwise 1x1 convolutions), as shown in Figure~\ref{fig:inc_blocks}. 
As such, when I3D inflates the convolutions to 3D, only some of the features contain temporal information.
However, by using  separable temporal convolution, we can add temporal information  to all 4 branches.
This improves the performance from $78.4\%$ to $78.9\%$ on \MK.
In the following sections, whenever we refer to an S3D model, we mean S3D with such configuration.
}

\label{sec:resultsSeparable}

Table~\ref{tab:FK}
compares the results of S3D and I3D on \FK. Table~\ref{tab:something} shows that S3D also outperforms I3D on the \Something\ dataset.
The results show that, despite a substantial compression in model size
(12.06M parameters for I3D reduced to 8.77M for S3D),
and a large speed-up ($107.9$ GFLOPS for I3D reduced to $66.38$ GFLOPS for 
S3D), the separable model is even more accurate
(top-1 accuracy improved from 71.1\% to 72.2\% for \FK, and from 45.8\% to 47.3\% for \Something).
We believe the gain in accuracy is because the
spatio-temporal
factorization reduces overfitting, in a way without sacrificing the expressiveness of the representation, as we find that simply reducing the parameters of the network does not help with the performance.

Note that we can apply this separable transformation to any place where 3D convolution is used;
thus this idea is orthogonal to the question of which layers should contain 3D convolution, which we discussed
in Section~\ref{sec:pyramids}.
We denote 
the separable version of the 
\pyramid\ models by \spyramid,
and the separable version of the 
\invpyramid\ models by \sinvpyramid,
thus 
giving us 4 families of models.

We plot the speed vs accuracy of these models
in Figure~\ref{fig:accuracyVsFlops}. We see that separable top-heavy models offer the best speed-accuracy trade-off.
In particular, the model in which we keep the top 2 layers as separable 3D convolutions,
and make the rest 2D convolutions,
seems to be a kind of ``sweet spot''.
We call this model ``Fast-S3D'',
since it is is 2.5 times more efficient than I3D
(43.47 vs 107.9 GFLOPS), and yet has comparable
accuracy (78.0\% vs 78.4\% on \MK).

\begin{figure}[htp]
\begin{center}
\includegraphics[height=59mm]{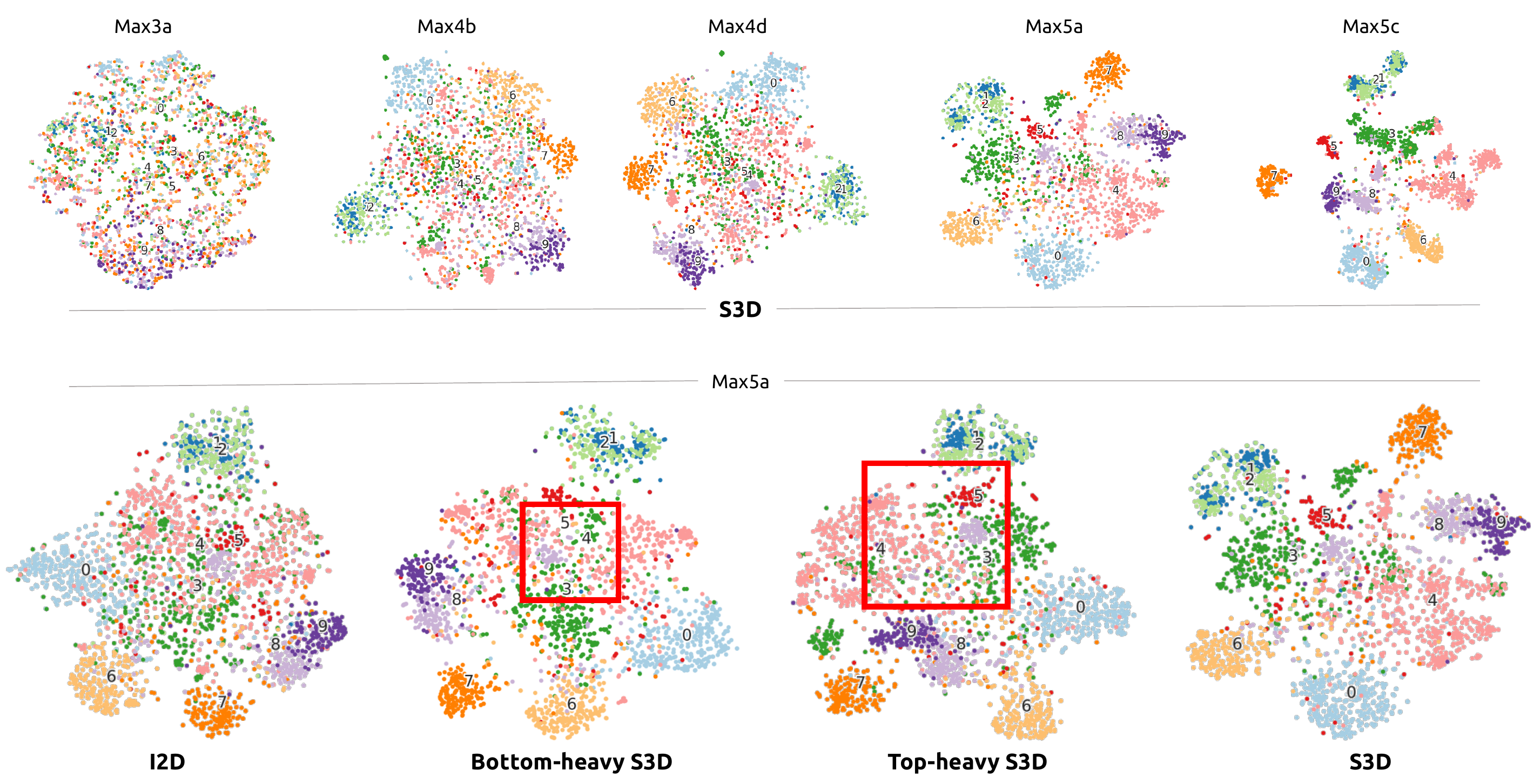}
\end{center}
\caption{tSNE projection of activation maps
derived from images in the \Something\ dataset.
Colors and numbers represent the 10 action groups defined in \cite{Something}.
The top row shows increasing semantic separation as we move to higher layers of S3D.
The bottom row shows activations at level Max5a for 4 different models.
We see that Top-Heavy-S3D has better semantic separation 
than Bottom-Heavy-S3D, especially for visually similar categories inside the red box.
}
\label{fig:tSNE}
\end{figure}

\subsection{tSNE analysis of the features}
\label{sec:tSNE}

Here we explore the spatiotemporal representations learned by different levels
of the S3D model 
on the \Something\ dataset,
using the tool of tSNE projection \cite{tSNE}. 
The behavior of the I3D models is very similar.
Instead of using samples from all 174 categories, we use a smaller vocabulary, namely the ``10 action groups'' defined in \cite{Something}.\footnote{
The labels are as follows.
0: Dropping [something], 
1: Moving [something] from right to left, 
2: Moving [something] from left to right, 
3: Picking [something], 
4:Putting [something],
5: Poking [something],
6: Tearing [something],
7: Pouring [something],
8: Holding [something],
9: Showing [something].
} %
We sample ~2,200 data points from the validation set. 
In Figure~\ref{fig:tSNE}, the top row shows representations learned by a S3D model,  at levels from Max3a to Max5c.
The class separation becomed increasingly clearer at higher levels. 

The bottom row shows representations learned at a certain feature level (Max5a), but across different models including I2D, Bottom-Heavy-S3D and Top-Heavy-S3D (both have a 2D-3D transition point at Max4b layer),
as well as a full S3D model. Comparing the bottom-heavy and top-heavy models, for subtle actions such as ``3: Picking'', ``4: Putting'' and ``5: Poking'' something, representations learned with a top-heavy model are more discriminative than that in a bottom-heavy model, thus leading to  better class separations with the tSNE projection (highlighted with the red box). A top-heavy model can learn features that are as good as those learned with a full 3D model, and significantly better than those from the 2D model, without much sacrifice in processing speed.
This observation further supports our hypothesis that temporal information modeling is most effective at top levels in the feature hierarchy for action classification tasks.

\begin{table}[!htp]
\begin{center}
\begin{tabular}{c|c|c|c|c}
\hline
Model  &  Top-1 (\%) & Top-5 (\%)  & Params (M) & FLOPS (G) \\
\hline
\Ithree  & 71.1 & 89.3 & $12.06$ & $107.89$ \\
\Sthree & 72.2 & 90.6 & $8.77$ & $66.38$ \\
\SG  & {\bf 74.7} & {\bf 93.4} & $11.56$ & $71.38$ \\
\hline
\end{tabular}
\end{center}
\caption{Effect of separable convolution and feature gating on the \FK\ validation set using RGB features.
}
\label{tab:FK}
\end{table}

\begin{table}[!htp]
\begin{center}
\begin{tabular}{c|c|c|c|c}
\hline
Model & Backbone & Val Top-1 (\%) & Val Top-5 (\%) & Test Top-1 (\%)\\
\hline
Pre-3D CNN + Avg~\cite{Something} & VGG-16 & - & - & 11.5 \\
Multi-scale TRN~\cite{zhou_trn} & Inception & 34.4 & 63.2 & 33.6 \\
\hline
\Itwo & Inception & 34.4 & 69.0 & - \\
\Ithree & Inception & 45.8 & 76.5 & - \\
\Sthree & Inception & 47.3 & 78.1 & - \\
\SG  & Inception & {\bf 48.2} & {\bf 78.7} & {\bf 42.0} \\
\hline
\end{tabular}
\end{center}
\caption{
Effect of separable convolution and feature gating on
the \Something\ validation and test sets  using RGB features.
}
\label{tab:something}
\end{table}

\subsection{Spatio-temporal feature gating}
\label{sec:gating}

In this section we further improve the accuracy of our model by using feature gating.
We start by considering the context feature gating
mechanism first used for video classification in \cite{miech2017learnable}. 
They consider an unstructured input feature vector $x \in \mathcal{R}^n$ (usually learned at final embedding layers close to the logit output), 
and produce an output feature vector $y \in \mathcal{R}^n$ as follows:
\[
y = \sigma(W x + b) \odot x
\]
where $\odot$ represents elementwise multiplication,
$W \in \mathcal{R}^{n \times n}$ is a weight matrix,
and $b \in \mathcal{R}^n$ is the bias term.
This mechanism allows the model to upweight certain 
dimensions of $x$ if the context model $\sigma(W x +b)$ predicts that they are important,
and to downweight irrelevant dimensions; this can be thought of as a ``self-attention'' mechanism.

We now extend this to feature tensors, with spatio-temporal structure.
Let $X \in \mathcal{R}^{T \times W \times H \times D}$ be the input tensor,
and let $Y$ be an output tensor of the same shape.
We replace the matrix product $W x$ with $W \mathrm{pool}(X)$,
where the pooling operation averages the dimensions of $X$ across space and time.
(We found that this worked better than  just averaging across space or just across time.)
We then compute $Y = \sigma(W \mathrm{pool}(X) + b) \odot X$,
where $\odot$ represents multiplication across the feature (channel) dimension,
(i.e., we replicate the attention map $\sigma(W \mathrm{pool}(X) + b)$ across space and time).

We can plug this gating module into any layer of the network.
We experimented with several options, and got the best results by applying it directly after
each of the $[k, 1, 1]$ temporal convolutions in the S3D network.
We call the final model (S3D with gating) \SG.
We see from Table~\ref{tab:FK} that this results in a healthy gain in accuracy 
compared to \Sthree\
on the \FK\ dataset (72.2\% top-1 to 74.7\%) at a very modest cost increase (66.38 GFLOPS to 71.38). Table~\ref{tab:something} shows that \SG\ also outperforms \Sthree\ and \Ithree\
on \Something.
We also significantly outperform the current state of the art method,
which is the Multi-scale TRN of \cite{zhou_trn},
improving top-1 accuracy from 33.6\% to 42.0\%.
\section{Generalization to other modalities, data and tasks}

In this section,  we evaluate the generality and robustness of the proposed \SG\ architecture by conducting transfer learning experiments on different input modalities, video datasets, and tasks. 

\subsection{Using optical flow features}
\label{sec:flow}

We first verify if \SG\ also works with optical flow inputs.
For these experiments, we follow the standard setup as described in~\cite{carreira2017quo} and extract optical flow features with the TV-L1 approach~\cite{tvl1}. We truncate the flow magnitude at $[-20, 20]$ and store them as encoded JPEG files. Other experiment settings are the same as the RGB experiments. From Table~\ref{tab:flow}, we can see that the improvement of \SG\ over I3D is consistent 
with the gain we saw with RGB inputs,
bringing the performance up from $63.91\%$ to $68.00\%$.
By ensembling
 the two streams of RGB and flow, we obtain
 a performance of 77.22\%,
 which is a 3\% boost over the I3D network when trained on the same data. We note that even though we focus on the speed-accuracy trade-offs in action classification network design, the performance is competitive compared with recent Kinetics Challenge winners and concurrent works; notably \cite{kinetics_winner} and \cite{wang2018non} use heavier backbone architectures (e.g. ResNet 101 has ~8.5x more FLOPS than our S3D-G architecture)
 
\begin{table*}[htp]
\begin{center}
\begin{tabular}{c|c|c|c|c|c}
\hline
Model & Inputs & Backbone & Pre-train & Top-1 (\%) & Top-5 (\%)\\
\hline
NL I3D \cite{wang2018non} & RGB & ResNet-101 & ImNet & {\bf 77.7} & {\bf 93.3} \\
SAN \cite{kinetics_winner} & RGB+Flow+Audio & Inception-ResNet-v2 & ImNet & {\bf 77.7} & {\bf 93.2} \\
TSN \cite{tsn_wang_eccv16} & RGB+Flow & Inception & ImNet & 73.9 & 91.1 \\
ARTNet \cite{ARTNet} & RGB+Flow & ResNet-18 & ImNet & 72.4 & 90.4\\
R(2+1)D \cite{Tran2018} & RGB+Flow & ResNet-34 & Sports-1M & 75.4 & 91.9 \\
\hline
I3D & Flow & Inception & ImNet & 63.9 & 85.0 \\
I3D & RGB & Inception & ImNet & 71.1 & 89.3 \\
I3D & RGB+Flow & Inception & ImNet & 74.1 & 91.6 \\
\hline
\SG & Flow & Inception & ImNet & 68.0 & 87.6 \\
\SG & RGB & Inception & ImNet & 74.7 & {\bf 93.4} \\
\SG & RGB+Flow & Inception & ImNet & {\bf 77.2}  & {\bf 93.0} \\
\hline
\end{tabular}
\end{center}
\caption{Benefits of using optical flow.
We report results
on the \FK\ validation set. 
We report I3D performance based on our implementation, as~\cite{carreira2017quo} only report results on the held-out test set (where they get a top-1 accuracy of 74.2\% using RGB+flow and ImNet pretraining).
}
\label{tab:flow}
\end{table*}

\subsection{Fine-tuning on other video classification datasets}
\label{sec:otherClassification}

Next we conduct transfer learning experiments from Kinetics to other video classification datasets,
namely HMDB-51~\cite{hmdb_kuehne2011} and UCF-101~\cite{ucf101}.
HMDB-51 contains around 7,000 videos spanning over 51 categories, while UCF-101 has 13,320 videos spanning over 101 categories. 
 Both datasets  consist of short video clips that are temporally trimmed, and contain 
 3 training and validation splits.
 We follow the standard setup as used in previous work and report average accuracy across all splits.

For our transfer learning experiments, we use the same setup as training on Kinetics, but change the number of GPUs to 8 and lower the learning rate to 0.01 for 6K steps, and 0.001 for another 2K steps.
For simplicity, we only use RGB (no optical flow).

Table~\ref{tab:cls-transfer} shows the results of this experiment. On UCF-101, our proposed \SG\ architecture,
which only uses Kinetics for pretraining,
outperforms I3D,
and matches R(2+1)D, both of which use largescale datasets (Kinetics and Sports-1M) for pretraining.
On HMDB-51, we outperform all previous methods published to date.

\begin{table}[pt!]
\small
\begin{center}
\begin{tabular}{c|c|c|c|c}
\hline
Model & Inputs & Pre-train & UCF-101 & HMDB-51 \\
\hline
P3D~\cite{Qiu_2017_ICCV} & RGB & Sports-1M & 88.6 & - \\
C3D~\cite{TranC3D} & RGB & Sports-1M & 82.3 & 51.6 \\
Res3D~\cite{TranR3D} & RGB & Sports-1M & 85.8 & 54.9 \\
ARTNet w/ TSN~\cite{ARTNet} & RGB & Kinetics & 94.3 & 70.9 \\
I3D~\cite{carreira2017quo} & RGB & ImNet+Kinetics & 95.6 & 74.8 \\
R(2+1)D \cite{Tran2018} & RGB & Kinetics & {\bf 96.8} & 74.5 \\
\hline
\SG	& RGB & ImNet+Kinetics & {\bf 96.8} & {\bf 75.9} \\
\hline
\end{tabular}
\end{center}
\caption{Results of various methods on action classification on the  UCF-101 and HMDB-51 datasets. 
All numbers are computed as the average accuracy across three splits.
}
\label{tab:cls-transfer}
\end{table}

\subsection{Spatio-temporal action detection in video}
\label{sec:detection}

Finally, we demonstrate the effectiveness of \SG\ on action detection tasks, where the inputs are video frames, and the outputs are bounding boxes associated with action labels on the frames.
Similar to the framework proposed in~\cite{peng2016multi}, we use the Faster-RCNN~\cite{ren2015faster} object detection algorithm to jointly perform person localization and action recognition. We use the same approach as described in~\cite{ava} to incorporate temporal context information via 3D networks. To be more specific, the model uses a 2D ResNet-50~\cite{resnet} network that takes the annotated keyframe (frame with box annotations) as input, and extract features for region proposal generation on the keyframe. We then use a 3D network (such as I3D or \SG) that takes the frames surrounding the keyframe as input, and extract feature maps which are then pooled for bounding box classification. The 2D region proposal network (RPN) and 3D action classification network are jointly trained end-to-end. Note that we extend the ROIPooling operation to handle 3D feature maps by simply pooling at the same spatial locations over all time steps.

We report performance on two widely adopted video action detection datasets: JHDMB~\cite{jhmdb} and UCF-101-24~\cite{ucf101}. JHMDB dataset is a subset of HMDB-51, it consists of 928 videos for 21 action categories, and each video clip contains 15 to 40 frames. UCF-101-24 is a subset of UCF-101 with 24 labels and 3207 videos; we use the cleaned bounding box annotations from~\cite{micro_tube2017}. We report performance using the standard frame-AP metric defined in~\cite{gkioxari2015}, which is computed as the average precision of action detection over all individual frames, at the intersection-over-union (IoU) threshold of 0.5. As commonly used by previous work, we report average performance over three splits of JHMDB and the first split for UCF-101-24.

Our implementation is based on the TensorFlow Object Detection API~\cite{huang2016speed}. We train Faster-RCNN with asynchronous SGD on 11 GPUs for 600K iterations. We fix the input resolution to be $320\times400$ pixels. For both training and validation, we fix the size of temporal context to 20 frames. All the other model parameters are set based on the recommended values from~\cite{huang2016speed}. The ResNet-50 networks are initialized with ImageNet pre-trained models, and I3D and \SG are pre-trained from Kinetics. We extract 3D feature maps from the ``\textit{Mixed 4e}'' layer which has a spatial stride of 16.

Table~\ref{tab:det-transfer} shows the comparison between I3D, \SG, and other state-of-the-art methods. We can see that both 3D networks outperform previous architectures by large margins, while \SG~is consistently better than I3D.

\begin{table}[!htp]
\begin{center}
\begin{tabular}{c|c|c|c}
\hline
Model & Inputs & JHMDB & UCF-101 \\
\hline
Gkioxari and Malik~\cite{gkioxari2015} &RGB+Flow & 36.2 & - \\
Weinzaepfel \etal~\cite{weinzaepfel2015} &RGB+Flow & 45.8 & 35.8 \\
Peng and Schmid~\cite{peng2016multi} & RGB+Flow & 58.5 & 65.7 \\
Kalogeiton \etal~\cite{kalogeiton17iccv} & RGB+Flow & 65.7 & 69.5 \\
\hline
Faster RCNN + I3D~\cite{ava} &RGB+Flow & 73.2 & 76.3 \\
Faster RCNN + \SG & RGB+Flow & {\bf 75.2} & {\bf 78.8} \\
\hline
\end{tabular}
\end{center}
\caption{Results of various methods on action detection in JHMDB and UCF101.
We report frame-mAP at IoU threshold of 0.5 on JHMDB (all splits) and UCF-101-24 (split 1) datasets.
}
\label{tab:det-transfer}
\end{table}

\section{Conclusion}
We show that we can significantly improve on the previous state of the art 3D CNN video classification model, known as I3D,
in terms of efficiency,
by combining 3 key ideas:
a top-heavy model design,
temporally separable convolution,
and spatio-temporal feature gating.
Our modifications are simple and can be applied to other architectures.
We hope this will boost performance on a variety of video understanding tasks.
\section*{Acknowledgment}
We would like to thank the authors
of \cite{kay2017kinetics} for the help on the Kinetics dataset and the baseline experiments,
especially Joao Carreira for many constructive discussions.
We also want to thank Abhinav Shrivastava, Jitendra
Malik, and Rahul Sukthankar for valuable feedbacks. S.X. is supported by Google. Z.T. is supported by NSF IIS-1618477 and NSF  IIS-1717431.

\bibliographystyle{splncs}
\bibliography{egbib}

\end{document}